\documentclass[lettersize,journal]{IEEEtran}
\usepackage{amsmath,amsfonts}
\usepackage{algorithmic}
\usepackage{algorithm}
\usepackage{array}
\usepackage[caption=false,font=normalsize,labelfont=sf,textfont=sf]{subfig}
\usepackage{textcomp}
\usepackage{stfloats}
\usepackage{url}
\usepackage{verbatim}
\usepackage{graphicx}
\usepackage{cite}
\usepackage{pifont}

\usepackage{framed,multirow}
\usepackage{colortbl}
\definecolor{newcolor}{rgb}{.8,.349,.1}
\definecolor{mygray}{gray}{.9}
\usepackage{booktabs}

\usepackage[normalem]{ulem} 

\begin{document}

\title{Cascaded Multi-Scale Attention for Enhanced Multi-Scale Feature Extraction and
Interaction with Low-Resolution Images}

\author{Xiangyong~Lu$^*$,~\IEEEmembership{}
	Masanori Suganuma$^{*\dag}$,~\IEEEmembership{}
	Takayuki Okatani$^{*\dag}$,~\IEEEmembership{}
	\IEEEcompsocitemizethanks{
		\IEEEcompsocthanksitem $*$Graduate School of Information Sciences, Tohoku University. 
		\IEEEcompsocthanksitem $^\dag$RIKEN Center for AIP
        }
	\thanks{}}

\markboth{IEEE TRANSACTIONS ON Multimedia,~Vol.~, No.~, Month~2025}%
{Shell \MakeLowercase{\textit{et al.}}: A Sample Article Using IEEEtran.cls for IEEE Journals}


\maketitle

\begin{abstract}
In real-world applications of image recognition tasks, such as human pose estimation, cameras often capture objects, like human bodies, at low resolutions. This scenario poses a challenge in extracting and leveraging multi-scale features, which is often essential for precise inference. To address this challenge, we propose a new attention mechanism, named cascaded multi-scale attention (CMSA), tailored for use in CNN-ViT hybrid architectures, to handle low-resolution inputs effectively. The design of CMSA enables the extraction and seamless integration of features across various scales without necessitating the downsampling of the input image or feature maps. This is achieved through a novel combination of grouped multi-head self-attention mechanisms with window-based local attention and cascaded fusion of multi-scale features over different scales.
This architecture allows for the effective handling of features across different scales, enhancing the model's ability to perform tasks such as human pose estimation, head pose estimation, and more with low-resolution images. Our experimental results show that the proposed method outperforms existing state-of-the-art methods in these areas with fewer parameters, showcasing its potential for broad application in real-world scenarios where capturing high-resolution images is not feasible. Code is available at https://github.com/xyongLu/CMSA. 
\end{abstract}


\section{Introduction}
\begin{figure}[t]
\centering
  \includegraphics[width=0.43\linewidth]{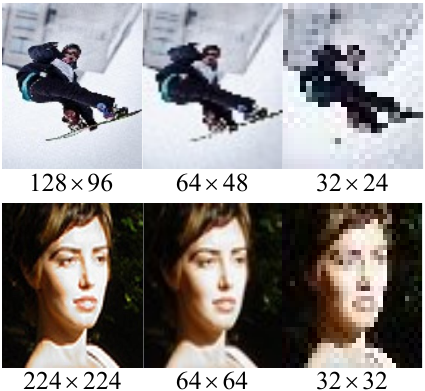}
  \raisebox{0mm}{\includegraphics[width=0.45\linewidth]{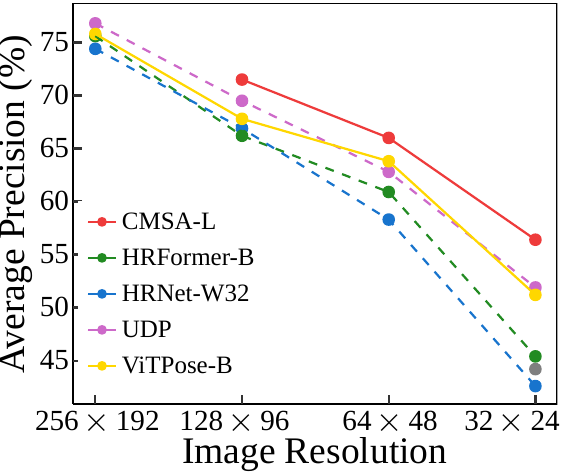}}
\caption{Left: Estimating pose from low-resolution images. Left-Upper: human pose estimation from COCO2017. Left-Lower: head pose estimation from AFLW2000. Right: resolution-accuracy trade-off for human pose estimation evaluated on COCO 2017 val dataset. }
\label{fig:tiny_imgs}
\end{figure}

\IEEEPARstart{I}{n} recent years, the application of deep learning for image recognition has significantly broadened its scope in the real world. At the forefront, there is the use under tougher conditions for image recognition. One of these is the application to low-resolution images. In this paper, we discuss how to maximize the estimation accuracy for several typical image recognition problems using images of a lower resolution than those primarily considered in previous studies, as illustrated in the left panel of Fig.~\ref{fig:tiny_imgs}.

There are several scenarios where handling low-resolution images is necessary. One is when ideal image capture conditions are not achievable, resulting in objects not being adequately resolved in images. For instance, this can occur when trying to estimate specific information, such as the pose of a person located at a relative distance from a fixed camera, like a surveillance camera. Another scenario involves conducting image recognition on low-performance edge devices. Being able to make estimations from lower-resolution images could lead to reductions in device costs and computational demands.

In this paper, we explore network architectures suitable for the above applications. Specifically, we build upon the recently evolved CNN-ViT hybrid models to design an architecture that can effectively handle low-resolution inputs. In particular, our goal is to perform multi-scale feature extraction from images and to facilitate appropriate interactions among these features.

Why do we aim to do this? First, multi-scale feature extraction combined with their interactions is an essential component for several tasks. The estimation of human poses mentioned earlier is a prime example of this necessity, as evidenced by previous studies \cite{ cheng2020higherhrnet, yuan2021hrformer, wang2022litepose}. While the level of necessity may differ, this strategy is fundamentally beneficial across a broad spectrum of image recognition tasks, including, for example, fine-grained object recognition \cite{ ibh2023tempose} and biomedical image segmentation \cite{ heidari2023hiformer}.

Second, low resolution complicates the management of multi-scale features. The conventional approach generates multi-scale features through the downsampling of feature maps \cite{sun2019hrnet, cheng2020higherhrnet, yuan2021hrformer}. This method is effective when starting with high-resolution images. However, with low-resolution inputs, repeated downsampling quickly diminishes the resolution of feature maps to the point where additional downsampling results in a significant loss of information.

To address this challenge, we investigate methods to manage multi-scale feature maps without downsampling, enabling interactions between features of varying scales. Specifically, we first divide the multi-heads in Multi-Head Self-Attention (MHSA) into several groups, each of which processes features of a different scale. Then, instead of downsampling, we adopt the window-based self-attention mechanism from the Swin Transformer \cite{liu2021swin} to extract a different scale feature in each group. To facilitate effective interactions among features of different scales, we transfer a feature map from one group to another in an order from lower to higher scales. The transferred feature is utilized in the attention calculations of the subsequent group. 

The approach to handling multi-scale features through grouping heads in the first part and the cascading calculation structure in the latter part are inspired by recent CNN-ViT hybrid models, SG-Former \cite{ren2023sgformer} and EfficientViT \cite{Liu2023efficientvit}, respectively. It is important to note that these existing models, while aiming to address the high computational demands of the ViT architecture and improve the trade-off between performance and computation, were not designed with low-resolution images or the extraction of multi-scale features plus their interactions in mind.

We present experimental results across four distinct tasks: human pose estimation, head pose estimation, image classification, and semantic segmentation, all utilizing low-resolution image inputs. For the first three tasks, the results demonstrate that our proposed method achieves superior performance to existing state-of-the-art methods tailored for these individual tasks; see the right panel of Fig.~\ref{fig:tiny_imgs}. Semantic segmentation experiments on the Cityscapes dataset provide preliminary evidence that our method has the potential to improve dense, pixel-level prediction under low-resolution settings. It is also noted that our method needs fewer parameters than the existing methods. 

\section{Related work}
\label{sec:ralatedwork}

\subsection{Extraction and Use of Multi-Scale Features}
Integrating features from multiple scales improves neural networks' ability to deeply understand complex visual scenes. 
HRNet \cite{sun2019hrnet, cheng2020higherhrnet} and HRFormer \cite{yuan2021hrformer} fuse the multi-scale representations from multiple branches to effectively generate reliable high-resolution representations with strong position sensitivity. 
CrossViT \cite{chen2021crossvit} learns multi-scale feature representations for image recognition by a dual-branch transformer and merges them multiple times to complement each other. 
FCT \cite{han2022FCTransformer} enhances cross-branch interactions in its two-branch transformer model to facilitate multi-scale representation learning.
TopFormer \cite{zhang2022topformer} produces scale-aware global features from tokens at various scales.
P2T \cite{wu2022p2t} and Shunted Transformer \cite{ren2022shunted} improve upon PVT \cite{wang2021pvt} by applying various downsampling rates for the keys and values across heads, rather than using identical spatial reduction for both. This allows each head to capture both coarse- and fine-grained details.
SG-Former \cite{ren2023sgformer} aggregates tokens at multiple scales and performs attention mechanisms to extract global and fine-grained information for visual tasks.
SMT \cite{lin2023scaleSMT} captures diverse spatial features through multiple convolutions with different kernel sizes and conducts scale-aware aggregation to enhance information interaction. 
SBCFormer  \cite{lu2024sbcformer} achieves efficient global and local feature fusion by introducing a bi-branch CNN-ViT hybrid module.
In existing methods, multi-scale representations are typically generated by downsampling the feature map, which can be problematic for low-resolution inputs, as it exacerbates the loss of detail. Our study mitigates this issue by enabling multi-scale feature extraction and interaction without the need for downsampling. 

\subsection{Dealing with Low-Resolution Images }

Low-resolution images often lack clarity in fine textures, edges, and small details, presenting significant challenges in image analysis. 
Some studies address this limitation through input enhancement or knowledge distillation. Haris \cite{haris2021lowobj} enhances low-resolution object detection by incorporating high-level vision objectives. DeriveNet \cite{singh2021derivenet} employs multi-resolution pyramid data augmentation to enrich feature representations, while CAL \cite{wang2022cal} uses Gaussian offset weighting to compensate for spatial displacements. FMD\cite{huang2022thinnet} proposes a feature distillation framework to transfer high-definition knowledge from a teacher model for improved low-resolution object recognition.

On the other hand, in order to reconstruct fine details from low-resolution images, recent Transformer-based super-resolution (SR) methods have emerged. 
ESRT \cite{lu2022transformer} combines a CNN-based backbone with a ViT-based backbone to recover a super-resolution image from its degraded low-resolution counterpart.
ESSAformer \cite{zhang2023essaformer} proposes an attention-embedded Transformer to enlarge the receptive field, allowing for better extraction of information from low-resolution hyperspectral images.
Steformer \cite{lin2023steformer} designs a bi-branch ViT-based structure to extract and integrate cross-view information from low-resolution stereo images.
However, SR methods typically rely on high-resolution supervision data, which is rarely accessible in real low-resolution settings. In contrast, our method preserves spatial details directly within each processing stage of CNN-ViT hybrids, operating exclusively on low-resolution inputs without high-resolution supervision.

\subsection{CNN-ViT Hybrids}

Hybrid models that combine convolutions and Vision Transformer (ViT) \cite{alexey2021ViT} are gaining interest for their ability to capture fine-grained local details while preserving global dependency. To enhance fine-grained information, some methods apply the attention mechanisms to sub-regions/tokens within feature maps. 
Swin \cite{liu2021swin} employs shifted windows in the self-attention mechanism to capture both local and global information. 
CSwin \cite{dong2022cswin} introduces a cross-shaped window to enhance the self-attention mechanism and further improve model capacity. 
NAT \cite{hassani2023neighborhood} proposes a scalable sliding window attention mechanism by localizing self-attention to the nearest neighboring pixels.
QFormer \cite{zhang2024qformer} extends the window-based attention mechanism within adaptive quadrangles for learning better feature representation.
With the demand for facilitating the aggregation of spatial information, many studies integrate convolutions within the attention mechanism.
MobileViT \cite{mehta2022mobilevit} and EfficientFormer \cite{li2022efficientformerV2} combine convolutions with Transformers to achieve efficient local and global feature extraction.
IRMB \cite{zhang2023IRMB} and SBCFormer \cite{lu2024sbcformer} introduce depth-wise separable convolutions into the self-attention mechanism to improve representational power. 
FastViT \cite{vasu2023fastvit} and RepViT \cite{wang2024repvit} employ structural parameterization techniques to optimize computational efficiency and model performance. 
Those methods primarily apply self-attention on reduced or locally grouped feature representations to balance efficiency and global context modeling. Under extremely low-resolution settings, such designs may restrict cross-scale spatial information exchange. CMSA addresses this by performing cascaded multi-scale attention to aggregate hierarchical multi-scale features.

\begin{figure*}[ht]
\begin{center}
   \includegraphics[width=1\linewidth]{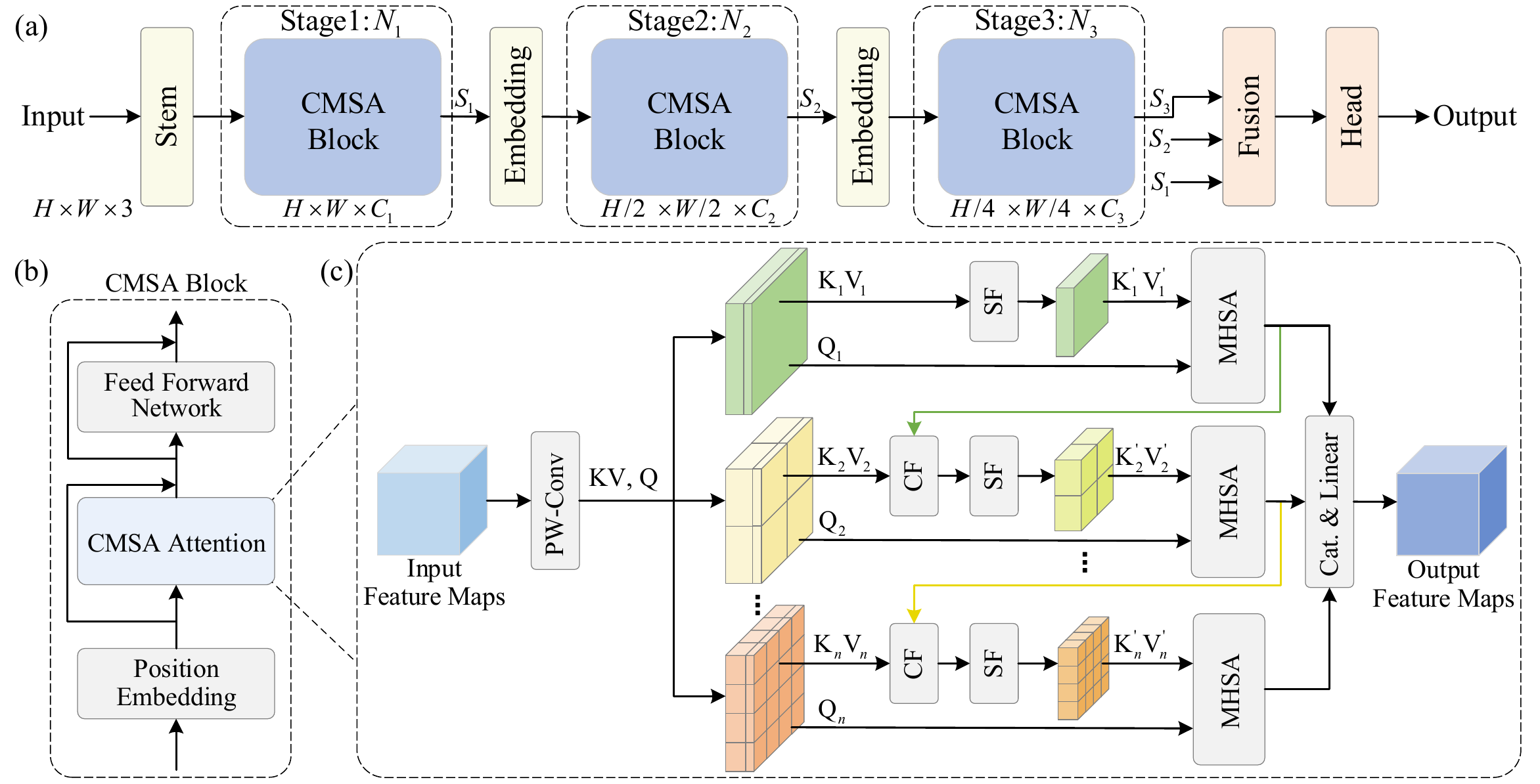}
\end{center}
   \caption{
   Design of the proposed model. (a) The overall architecture. (b) Design of the basic blocks. (c) Proposed cascaded multi-scale attention (CMSA). See texts for details.}
\label{fig:cmsa}
\end{figure*}

\section{Attention Mechanism for Low-Resolution Inputs}
\label{sec:method}

\subsection{Motivation}

In various tasks including pose estimation, it's widely acknowledged that enabling multi-scale feature extraction and interaction between different scale features is crucial for achieving high-precision inference. In existing architectures, the multi-scale space is typically generated by simply downsampling the feature maps. However, this method poses a challenge when the input image resolution is low, as downsampling leads to a substantial reduction in the resolution of the resulting feature maps, leading to a considerable loss of information\footnote{Regardless of the input resolution, stage-wise integration of spatial information via downsampling, which is fundamental to CNN-ViT hybrids, remains effective. The discussion above addresses a different aspect, focusing on the multi-scale feature processing {\em within} each processing stage in CNN-ViT hybrids.}.

\begin{figure}[ht]
\begin{center}
   \includegraphics[width=0.8\linewidth]{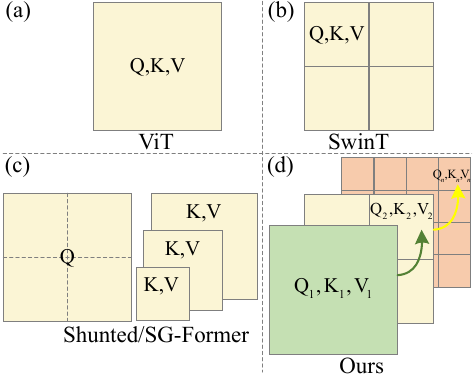}
\end{center}
   \caption{
   Comparison of receptive fields and feature fusion across different attention mechanisms. (a) ViT uses single-scale global self-attention with a fixed receptive field. (b) SwinT adopts uniform local-window attention, restricting interactions to fixed-size windows. (c) Shunted/SG-Former applies hybrid-scale attention through multi-scale token aggregation, reducing spatial resolution. (d) Our CMSA partitions attention heads into $n$ groups with distinct window sizes and processes them sequentially, enabling multi-scale receptive fields and directional cross-scale feature fusion at full resolution.
   }
\label{fig:cmsa_fea}
\end{figure}

To overcome this challenge, our study aims to devise an effective method to support multi-scale feature extraction and interaction without downsampling. We approach this goal by enhancing the attention mechanism in CNN-ViT hybrids.
As illustrated in Fig.~\ref{fig:cmsa}c, our CMSA partitions attention heads into groups with distinct window sizes and processes them sequentially, forming multi-scale receptive fields and enabling directional cross-scale feature fusion. In contrast to single-scale ViT attention \cite{alexey2021ViT}, SwinT's uniform local-window attention \cite{liu2021swin}, and the token-merging hybrid-scale attention of Shunted/SG-Former \cite{ren2022shunted, ren2023sgformer} shown in Fig. \ref{fig:cmsa_fea}, this design avoids spatial resolution reduction during feature fusion, thereby preserving fine spatial details under low-resolution inputs.

\subsection{Basic Ideas: Cascaded Multi-Scale Attention (CMSA) }

There are two primary ideas. The first involves directing each head of the multi-head self-attention mechanism to handle features from different scales. While conventional multi-head self-attention mechanisms already possess sufficient flexibility, they may not inherently fulfill this role. Contrarily, previous research has shown that attention maps tend to exhibit similarity between heads, leading to redundancy \cite{ bolya2022reddncyeffic, chen2022reddncydiverse}.

To address this, we propose organizing the multi-heads into groups and implementing SwinT's local window attention \cite{liu2021swin} within each group; see Fig.~\ref{fig:cmsa}c. We vary the window size for each group. This strategy aims to force different attention heads to effectively manage features of different scales. Our approach draws inspiration from the hybrid-scale attention (HSA) in SG-Former \cite{ren2023sgformer} and Shunted Transformer\cite{ren2022shunted}. However, HSA does not use SwinT's local attention and instead adjusts key and value through token merging that reduces spatial resolution at different scales, as visualized in Fig.~\ref{fig:cmsa_fea}c \footnote{While we initially experimented with the hybrid-scale attention mechanism, we found that our method based on local attention works much better.}.

Our second idea involves implementing a new mechanism to enhance interaction among features of different scales across various groups. This entails organizing the groups in a specific order and transferring features obtained from local attention within one group to the next. These features are subsequently integrated into the attention computation within each group, resulting in an architecture that operates in a cascade-style fashion, as depicted in Fig.~\ref{fig:cmsa}c.

This approach draws inspiration from EfficientViT \cite{Liu2023efficientvit}. It divides multi-head attention into groups and performs cascaded information integration, and our method follows the same principle. However, in EfficientViT, processing within groups primarily employs standard self-attention, and the scale of feature maps is the same across all groups. Its aim is to improve the trade-off between computational cost and expressive power. In contrast, our approach emphasizes fostering interactions between features of different scales. We also design a new method for integrating information between groups to achieve this goal.

\subsection{Details of CMSA } 
\label{sec:attn}

We will now describe the cascaded multi-scale attention mechanism; see Fig.~\ref{fig:cmsa}c. It adopts a cascaded multi-scale feature integration strategy, implemented within a single attention layer. Let $\mathbf{X}$ be the input to our cascaded multi-scale attention mechanism; $\mathbf{X}$ is a feature map with the size ${H \times W \times C}$. We first apply point-wise convolutions (PW-Conv) to $\mathbf{X}$ to obtain  
$\mathbf{Q}\in \mathbb{R}^{H \times W \times D}$, $\mathbf{K}\in \mathbb{R}^{H \times W \times D}$, and $\mathbf{V}\in \mathbb{R}^{H \times W \times D}$. 

We employ multi-head self-attention (MHSA), similar to the standard Transformer architecture. Specifically, $\mathbf{Q}$, $\mathbf{K}$, and $\mathbf{V}$ are split (and reshaped) into $\tilde{\mathbf{Q}}_i$, $\tilde{\mathbf{K}}_i$, and $\tilde{\mathbf{V}}_i\in\mathbb{R}^{HW\times d_i}$ ($i=1,\ldots,h$), where $d_i$ is the feature dimension in this head. Then each head $i$ conducts a self-attention computation as $\mathrm{softmax}(\tilde{\mathbf{Q}}_i\cdot \tilde{\mathbf{K}}_i^\top/\sqrt{d_i})\cdot \tilde{\mathbf{V}}_i$. 

To deal with multi-scale features, we employ a group structure, as mentioned earlier. 
Specifically, we divide the heads into $n$ groups, enabling each group to extract and foster features with different scales. Thus, $\mathbf{Q}$, $\mathbf{K}$, and $\mathbf{V}$ are segmented in their channel dimensions to yield query, key, and value for each group as
$\mathbf{K}_1$,\ldots,$\mathbf{K}_n$, $\mathbf{Q}_1$,\ldots,$\mathbf{Q}_n$, and 
$\mathbf{V}_1$,\ldots,$\mathbf{V}_n$; $\mathbf{Q}_k$, $\mathbf{K}_k$, $\mathbf{V}_k\in\mathbb{R}^{H\times W\times d_k}$. 

We employ SwinT's local window attention in each group. Let the index of groups be $k=1,\ldots,n$. For the $k$-th group, we set the local window size to $s_k\times t_k$. Thus, the $k$-th group has $H/s_k \times W/t_k$ local windows, each with $s_k\times t_k$ size. We set $(s_1, t_1)=(H, W)$, meaning that the first head group ($k=1$) extracts global features as the window size is the same as the input feature map (i.e., $H\times W$). The subsequent groups, with $s_k<H$ and $t_k<W$,  extract local features with the specified window size $s_k\times t_k$. 

To enable different scale features to interact with each other, we incorporate the cascaded structure of feature propagation and integration over the head groups, as explained above. 
Specifically, the $k$-th stream receives $\mathbf{Q}_k$, $\mathbf{K}_k$, and $\mathbf{V}_k$ as inputs and outputs an updated feature map $\mathbf{X}_k'\in\mathbb{R}^{H\times W\times d_k}$ at its end. This output $\mathbf{X}_k'$ is transferred to the intermediate section of the next (i.e., $(k+1)$-th) stream, contributing to the computation of the stream's output $\mathbf{X}_{k+1}'$.

Specifically, the $k$-th group, starting with $\mathbf{Q}_k$, $\mathbf{K}_k$, and $\mathbf{V}_k$, receives the output $\mathbf{X}'_{k-1}$ from the lower stream and updates $\mathbf{K}_k$ and $\mathbf{V}_k$ as
\begin{equation}
  \label{eq:kv}
  \mathbf{K}'_k, \mathbf{V}'_k = {\rm SF}({\rm CF}({\rm Concat}(\mathbf{K}_k, \mathbf{V}_k, \mathbf{X}'_{k-1}))),
\end{equation}
where Concat indicates the concatenation of $\mathbf{K}_k$, $\mathbf{V}_k$, and $\mathbf{X}'_{k-1}$ in their channel dimension, and the attention computation within each group is fully parallelizable;
CF (channel fusion) is a point-wise convolution, aiming to make its input exchange information in the channel dimension; on the other hand, SF (spatial fusion) aims at spatial exchange of information in the feature map. Specifically, it is implemented as a series of a depth-wise convolution with a $3\times 3$ kernel, a GeLU activation function, and a point-wise convolution\footnote{An optional average pooling with a $2\times 2$ kernel to reduce the size of $K$ and $V$.}.  The combination of SF and CF allows efficient multi-scale feature integration along both spatial and channel dimensions, supporting cascaded feature aggregation across groups. For the first stream with $k=1$, it does not have a lower stream and thus $\mathbf{K}_1'$ and $\mathbf{V}_1'$ are computed as $\mathrm{SF}(\mathrm{Concat}(\mathbf{K}_1,\mathbf{V}_1))$.

Then, the SwinT’s local window attention with a window size $s_k\times t_k$ is computed between the query $\mathbf{Q}_k$ and the updated key $\mathbf{K}_k'$ and is applied to the updated value $\mathbf{V}_k'$, yielding the feature map $\mathbf{X}_k'$. Note that each group has multiple heads in each of which local, multi-head attention is computed independently. 
Finally, all the outputs from all the streams are aggregated to yield the output $\mathbf{X}''$ of the attention mechanism as
\begin{equation}
  \label{eq:proj_linear}
  \mathbf{X}'' = {\rm Linear}({\rm Concat}(\mathbf{X}'_1,..., \mathbf{X}'_n)),
\end{equation}
where $\mathrm{Linear}$ is a linear transformation implemented as a linear layer with learnable weights. 

\paragraph{Optional Setting of Attention}
So far, we have considered the standard setting of attention where the query, key, and value all share the same dimensions. Prior studies \cite{graham2021levit,wang2022pvtv2,guo2022cmt} demonstrate that making the key and value spatially smaller (or equivalently, using fewer tokens) than the query can enhance computational efficiency with the minimum sacrifice of inference accuracy. Specifically, we adjust the convolution in the spatial fusion (SF) module to halve the spatial dimensions of the key and value compared to the query, while doubling the channel size of the value relative to both the query and key. In our experiments, we employ this method to achieve a better trade-off of accuracy and computational cost. 

\begin{table*}[ht]
\newcommand{\tabincell}[2]{\begin{tabular}{@{}#1@{}}#2\end{tabular}}
\renewcommand{\arraystretch}{1.} 
\centering
\caption{CMSA variants with different model sizes. \quad \quad \quad \quad   }
\begin{tabular}{c|c|c|c c c|c c c|c c c}
\toprule
\multirow{2}*{\textbf{Stage}} &  \multirow{2}*{\textbf{\shortstack{ Output \\Size} }} &
\multicolumn{1}{c|}{\multirow{2}*{\textbf{Block}}} & \multicolumn{9}{c}{\textbf{ CMSA }} \\ \cline{4-12}
& & & \multicolumn{3}{c|}{\textbf{S}} & \multicolumn{3}{c|}{\textbf{B}} & \multicolumn{3}{c}{\textbf{L}} \\
 \hline \hline
\multirow{2}*{Stem} &  \multirow{2}*{ 32$\times$24} & \multicolumn{1}{c|}{\multirow{2}*{\shortstack{Patch \\Embed.}}}  &  \multicolumn{9}{c}{$\times$ 2 ($k=3\times 3$)} \\ 
\cline{4-12} 
& & & \multicolumn{3}{c|}{dim. 96} & \multicolumn{3}{c|}{dim. 128} & \multicolumn{3}{c}{dim. 128} \\ 
\hline 
\multirow{6}*{1} &  \multirow{6}*{ 32$\times$24} & \multicolumn{1}{c|}{\multirow{6}*{\shortstack{ CMSA \\Block}}} &  \multicolumn{3}{c|}{$\times$ 2} & \multicolumn{3}{c|}{$\times$ 2} & \multicolumn{3}{c}{$\times$ 2} \\ 
\cline{4-12}
& & & \multicolumn{3}{c|}{$n=3$ groups} & \multicolumn{3}{c|}{$n=3$ groups} & \multicolumn{3}{c}{$n=3$ groups} \\ 
\cline{4-12}
& & & \multicolumn{1}{c|}{\tabincell{c}{$s_1$=32 \\$t_1$=24 \\ $d_1$=32 \\ $head$=2}} & \multicolumn{1}{c|}{\tabincell{c}{$s_2$=16 \\$t_2$=12 \\ $d_2$=32 \\ $head$=2}} & \multicolumn{1}{c|}{\tabincell{c}{$s_3$=8 \\$t_3$=6 \\ $d_3$=16 \\ $head$=1}} &
\multicolumn{1}{c|}{\tabincell{c}{$s_1$=32 \\$t_1$=24 \\ $d_1$=16 \\ $head$=1}} & \multicolumn{1}{c|}{\tabincell{c}{$s_2$=16 \\$t_2$=12 \\ $d_2$=32 \\ $head$=2}} & \multicolumn{1}{c|}{\tabincell{c}{$s_3$=8 \\$t_3$=6 \\ $d_3$=16 \\ $head$=1}} & 
\multicolumn{1}{c|}{\tabincell{c}{$s_1$=32 \\$t_1$=24 \\ $d_1$=16 \\ $head$=1}} & \multicolumn{1}{c|}{\tabincell{c}{$s_2$=16 \\$t_2$=12 \\ $d_2$=32 \\ $head$=2}} & \multicolumn{1}{c}{\tabincell{c}{$s_3$=8 \\$t_3$=6 \\ $d_3$=16 \\ $head$=1}} \\ 
\hline

\multirow{8}*{2} & \multirow{8}*{ 16$\times$12}& \multirow{2}*{\shortstack{Patch \\Embed.}}&\multicolumn{9}{c}{$\times$ 1 ($k=3\times 3$)} \\
\cline{4-12}
& & & \multicolumn{3}{c|}{dim. 160} & \multicolumn{3}{c|}{dim. 192} & \multicolumn{3}{c}{dim. 256} \\
\cline{3-12}
& & \multicolumn{1}{c|}{\multirow{6}*{\shortstack{ CMSA \\Block}}} & \multicolumn{3}{c|}{$\times$ 4} & \multicolumn{3}{c|}{$\times$ 4} & \multicolumn{3}{c}{$\times$ 4} \\ 
\cline{4-12}
& & & \multicolumn{3}{c|}{$n=2$ groups} & \multicolumn{3}{c|}{$n=2$ groups} & \multicolumn{3}{c}{$n=2$ groups} \\ 
\cline{4-12}
& & &  \multicolumn{1}{c}{\tabincell{c|}{$s_1$=16 \\$t_1$=12 \\ $d_1$=48 \\ $head$=3}} & \multicolumn{1}{c}{\tabincell{c|}{$s_2$=8 \\$t_2$=6 \\ $d_2$=48 \\ $head$=3}} & & \multicolumn{1}{c|}{\tabincell{c}{$s_1$=16 \\$t_1$=12 \\ $d_1$=48 \\ $head$=3}} & \multicolumn{1}{c|}{\tabincell{c}{$s_2$=8 \\$t_2$=6 \\ $d_2$=48 \\ $head$=3}} & &
\multicolumn{1}{c|}{\tabincell{c}{$s_1$=16 \\$t_1$=12 \\ $d_1$=48 \\ $head$=3}} & \multicolumn{1}{c|}{\tabincell{c}{$s_2$=8 \\$t_2$=6 \\ $d_2$=48 \\ $head$=3}}\\ 
\hline
\multirow{8}*{3} & \multirow{8}*{ 8$\times$6}& \multirow{2}*{\shortstack{Patch \\Embed.}}&\multicolumn{9}{c}{$\times$ 1 ($k=3\times 3$)} \\
\cline{4-12}
& & & \multicolumn{3}{c|}{dim. 224} & \multicolumn{3}{c|}{dim. 256} & \multicolumn{3}{c}{dim. 320} \\
\cline{3-12}
& & \multicolumn{1}{c|}{\multirow{6}*{\shortstack{ CMSA \\Block}}} &  \multicolumn{3}{c|}{$\times$ 3} & \multicolumn{3}{c|}{$\times$ 3} & \multicolumn{3}{c}{$\times$ 3} \\ 
\cline{4-12}
& & & \multicolumn{3}{c|}{$n=2$ groups} & \multicolumn{3}{c|}{$n=2$ groups} & \multicolumn{3}{c}{$n=2$ groups} \\ 
\cline{4-12}
& & &  \multicolumn{1}{c|}{\tabincell{c}{$s_1$=8$\,$ \\$t_1$=6 \\ $d_1$=64 \\ $head$=4}} & \multicolumn{1}{c|}{\tabincell{c}{$s_2$=8$\,$ \\$t_2$=6 \\ $d_2$=64 \\ $head$=4}} & & \multicolumn{1}{c|}{\tabincell{c}{$s_1$=8$\,$ \\$t_1$=6 \\ $d_1$=64 \\ $head$=4}} & \multicolumn{1}{c|}{\tabincell{c}{$s_2$=8$\,$ \\$t_2$=6 \\ $d_2$=80 \\ $head$=5}} & & \multicolumn{1}{c}{\tabincell{c|}{$s_1$=8$\,$ \\$t_1$=6 \\ $d_1$=64 \\ $head$=4}} & \multicolumn{1}{c}{\tabincell{c|}{$s_2$=8$\,$ \\$t_2$=6 \\ $d_2$=64 \\ $head$=4}} \\ 
\hline
\bottomrule
\end{tabular}
\label{tab:arch_details}
\end{table*}
\subsection{Design of a Whole Network}

\subsubsection{Overall Design} 
We adopt a hierarchical pyramidal architecture, in line with recent CNN-ViT hybrids; see Fig.~\ref {fig:cmsa}a. The architecture starts with a `Stem' block, followed by a sequence of stages. Each stage comprises multiple blocks that include the cascaded multi-scale attention mechanism explained above. A patch embedding layer is placed between stages to reduce the spatial resolution of feature maps. 

The process begins with an input image of size $H\times W\times 3$, which is converted by the Stem into a feature map of dimensions $H\times W\times C$. With each subsequent stage, the resolution is halved, leading to feature map sizes of $H\times W$, $H/2\times W/2$, and $H/4\times W/4$, respectively. In the final stage, feature maps from each block are merged and funneled through a task-specific head component, resulting in the final output.

\subsubsection{Design of a Block}
\label{sec:block}

As shown in Fig.~\ref{fig:cmsa}b, a single block consists of three components: position embedding, the cascaded multi-scale attention, and a feed-forward network.

For position embedding, we employ conditional positional embedding via depth-wise convolution. 
\begin{equation}
\mathbf{X}_\mathrm{PE}=\mathrm{PE}(\mathbf{X}_\mathrm{in})
\end{equation}
We choose $3\times 3$ depth-wise convolution for $\mathrm{PE}$. We incorporate a train-time overparametrization strategy \cite{chu2021conditionalPE,vasu2023fastvit},  aiming at a better trade-off between inference accuracy and computational cost. Specifically, our network features three parallel branches in this section during the training phase: a $3\times 3$ depth-wise convolution, a point-wise convolution, and batch normalization. These branches are reparameterized into a single depth-wise convolution at the testing phase. We apply GeLU activation after the depth-wise convolution.

The resulting feature map is fed into the aforementioned cascaded multi-scale attention mechanism (CMAttn) with one residual connection as
\begin{equation}
  \label{eq:rec_attn}
  \mathbf{X}_\mathrm{attn} =   \mathrm{CMAttn}(\mathrm{BN}(\mathbf{X}_\mathrm{PE})) + \mathbf{X}_\mathrm{PE},
\end{equation}
where BN denotes Batch Normalization. Its output is fed into a feed-forward network (FFN), following the standard design of ViTs \cite{alexey2021ViT} as
\begin{equation}
  \label{eq:ffn}
  \mathbf{X}_\mathrm{out} =   \mathrm{FFN}(\mathrm {LN}(\mathbf{X}_\mathrm{attn})) + \mathbf{X}_\mathrm{attn},
\end{equation}
where LN denotes Layer Normalization. For the FNN, we employ a two-layer MLP with 
a GeLU activation function at its hidden layer. 

The patch embedding layer comprises reparameterizable depth-wise and point-wise convolution layers, which use train-time overparameterization, and these additional branches are removed by structural reparameterization at inference \cite{ding2021paramDiversebb}.

\subsubsection{Other Components}
The Stem block and patch embedding layers between stages are implemented as a convolutional layer. 
Specifically, we adopt a single pair consisting of a depth-wise convolution followed by a point-wise convolution for each, aiming to boost computational efficiency. For implementation, we adopt a train-time overparametrization strategy \cite{vasu2023mobileone}, similar to the one used for positional embedding, to achieve a favorable balance between accuracy and efficiency. 

\begin{table*}[t]
\centering
\caption{Results on bottom-up human pose estimation with the COCO 2017 val dataset. All FLOPs are calculated at the respective single-scale. 
The speed of all methods is recorded on a single RTX 3090 GPU with a batch size of 32. $^\dagger$ means the results reported in the respective papers. $*$ indicates models trained with a different learning rate scheduler.}
\smallskip
\begin{tabular}{c | c| c | c | c | c | c c c c c c }
\toprule
\multirow{2}*{Model} &\multirow{2}*{Backbone}  & Input & Params  &FLOPs & Speed & \multicolumn{6}{c}{COCO 2017 val} \\
\cline{7-12} & & Resolution& (M) &(G)& (fps) &\textbf{AP}$\uparrow$& AP$^{50}$  & AP$^{75}$ & AP$^{M}$ & AP$^{L}$ & AR$\uparrow$ \\
 \hline
SimpleBaseline$^\dagger$ \cite{xiao2018simple} & ResNet-50&$128\times 96$& 34.0& 2.3&697.1 &59.3& 85.5& 67.4& 57.8& 63.8& 66.6 \\
HRNet-W32$^\dagger$ \cite{sun2019hrnet} & HRNet-W32 &$128\times 96$ & 28.5& 1.9&318.1 &66.9&  88.7& 76.3& 64.6& 72.3& 73.7\\
HRNet-W48$^\dagger$ \cite{sun2019hrnet}& HRNet-W48 &$128\times 96$& 63.6 &3.9&316.3 & 68.0& 88.9& 77.4& 65.7& 73.7& 74.7 \\
DARK$^\dagger$ \cite{zhang2020dark}& HRNet-W32 &$128\times 96$& 28.5&1.9& 257.4& 70.7& 88.9& 78.4& 67.9& 76.6& \textbf{76.7}  \\
UDP \cite{huang2020udp}& HRNet-W32 &$128\times 96$& 28.6&1.9&307.8 & 69.5& 91.3& 77.7& 67.9& 72.2& 73.7  \\
HRFormer-B  \cite{yuan2021hrformer}& HRFormer-B &$128\times 96$& 43.2& 3.7 &176.6 & 66.2& 88.2& 76.1& 64.1& 71.8& 73.0 \\
ViTPose-B  \cite{xu2022vitpose}& ViT-B &$128\times 96$& 86.0& 17.0 &190.6 &67.8& 88.3& 76.2& 65.9& 73.3& 74.6 \\
PCT-B \cite{geng2023pct}& Swin-B &$128\times 96$ &87.0 &11.6 & 137.2 &66.4 &90.5 &75.7 &65.1 &68.7 &70.1  \\
\rowcolor{mygray}
\textbf{CMSA-B}& CMSA-B &$128\times 96$& 5.6& 4.1 &761.0 &70.3& 91.3& 77.8& 67.9& \textbf{77.4} & 74.6 \\
\rowcolor{mygray}
\textbf{CMSA-L}& CMSA-L &$128\times 96$& 7.3& 5.1 &728.2 &\textbf{71.8}& \textbf{92.4}& \textbf{80.6}& \textbf{69.8}& 75.0& 75.7\\
\hline
HRNet-W32 \cite{sun2019hrnet} & HRNet-W32 &$64\times 48$ & 28.5& 1.9 & 321.9 &58.3& 86.2& 67.1& 57.4& 62.3& 66.2  \\ 
CAL $^\dagger$ \cite{wang2022cal}& HRNet-W48  &$64\times 48$ & 110.3& 5.3&- &61.5& 88.1& 68.7& 60.7& 63.5& 66.3  \\
UDP \cite{huang2020udp}& HRNet-W32 &$64\times 48$& 28.6&1.9 &311.4 &62.8& 86.4& 70.9& 61.3& 67.8& 70.4 \\
HRFormer-B  \cite{yuan2021hrformer}& HRFormer-B &$64\times 48$& 43.2&3.7 & 176.0&60.9& 87.1& 70.6& 60.0& 65.2& 68.7 \\
ViTPose-B  \cite{xu2022vitpose}& ViT-B &$64\times 48$& 86.0&16.9 & 191.9&63.8& 86.5& 72.6& 62.5& 68.7& 71.4  \\
PCT-B \cite{geng2023pct}& Swin-B &$64\times 48$ &87.0 &3.9 & 174.7  &62.0& 90.5 &71.9 &61.4 &63.8 &66.7  \\
\rowcolor{mygray}
\textbf{CMSA-B}$^*$& CMSA-B & $64\times 48$& 5.6& 3.7&809.7 &64.7& 87.6& 72.3& 62.9& 69.7 & 71.7 \\
\rowcolor{mygray}
\textbf{CMSA-L}$^*$& CMSA-L & $64\times 48$& 7.3& 4.6 &785.8 &65.9& 88.1& 74.5& \textbf{64.0}& \textbf{70.2}& \textbf{72.7} \\
\rowcolor{mygray}
\textbf{CMSA-B}& CMSA-B & $64\times 48$& 5.6& 3.7&809.7 &65.2& 91.4& 75.3& 63.0& 68.7 & 69.3 \\
\rowcolor{mygray}
\textbf{CMSA-L}& CMSA-L & $64\times 48$& 7.3& 4.6 &785.8 &\textbf{66.0}& \textbf{91.5}& \textbf{76.4}& 63.6& 69.5& 70.0 \\
\hline
HRNet-W32  \cite{sun2019hrnet}& HRNet-W32 &$32\times 24$ & 28.5&1.9 &326.8 &42.6& 79.6& 41.5& 43.4& 44.6& 52.7  \\
UDP \cite{huang2020udp}& HRNet-W32 &$32\times 24$& 28.6&1.9 & 312.5 &51.9& 81.7& 56.8& 51.5& 55.9& \textbf{61.4} \\
HRFormer-B  \cite{yuan2021hrformer}& HRFormer-B &$32\times 24$& 43.2&3.7 &177.5 &45.4& 81.8& 46.2& 45.9& 48.0& 55.3  \\
ViTPose-B  \cite{xu2022vitpose}& ViT-B &$32\times 24$& 86.0&16.9 & 193.5 &51.2& 81.0& 56.1& 50.7& 55.0& 60.4\\
PCT-B \cite{geng2023pct}& Swin-B &$32\times 24$ &87.0&3.9 & 178.6 &44.4 &84.6 &41.8 &45.3 &43.7 &50.7  \\
\rowcolor{mygray}
\textbf{CMSA-S}& CMSA-S & $32\times 24$& 4.1& 0.7& 1078.8 &52.3& 84.0& 57.3& 51.6& 53.9& 57.2 \\
\rowcolor{mygray}
\textbf{CMSA-B}& CMSA-B & $32\times 24$& 5.6& 0.9 &1005.4 &53.5& 84.9& 58.1& 53.2& 54.7& 58.5\\
\rowcolor{mygray}
\textbf{CMSA-L}& CMSA-L & $32\times 24$& 7.3& 1.1 &986.9 &\textbf{56.4}& \textbf{87.0}& \textbf{62.6}& \textbf{55.6}& \textbf{58.2}& 61.1\\
\bottomrule
\end{tabular}
\label{table:hum_pose}
\end{table*}

\section{Experimental Results}
\label{sec:experitments}
We evaluate the effectiveness of our proposed method and compare it with existing models through experiments on three distinct tasks: human pose estimation, head pose estimation, and small-image classification using CIFAR-10/100 datasets. 

\subsection{Human Pose Estimation}
\subsubsection{Dataset}
We use the COCO 2017 dataset  \cite{lin2014micrococo}, which includes 118,000 training and 5,000 validation, and 20,000 test images. Following the common bottom-up setting\footnote{We use the official Github repository of HRNet \cite{sun2019hrnet} as a framework for the experiments: https://github.com/leoxiaobin/deep-high-resolution-net.pytorch. }, individual persons' bounding boxes are provided. Each person is labeled with 17 keypoints, and the models are tasked with predicting these keypoints' coordinates based on the cropped and resized image from its bounding box in the input image. The dataset contains 250,000 individuals with heights ranging from 32 to over 128 pixels. These images are resized to a standard size before being fed into a model, and we adjust this size to manipulate the input resolution. 

\begin{figure}[ht]
\begin{center}
   \includegraphics[width=0.99\linewidth]{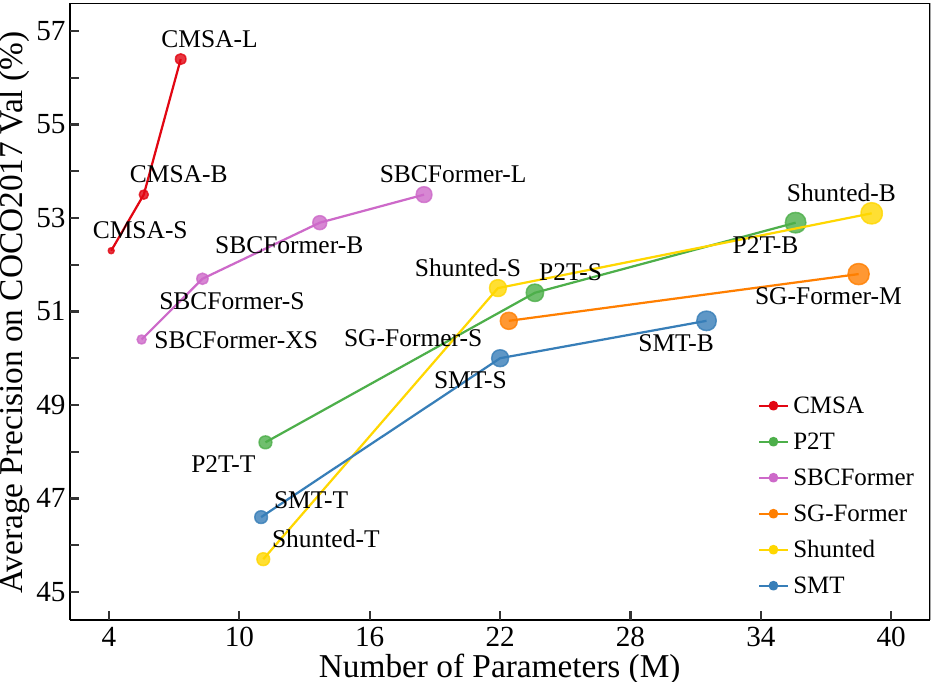}
   \end{center}
   \caption{Average Precision vs. Parameters. All models are trained for human pose estimation with input size $32\times 24$.}
\label{fig:ap_param_pose}
\end{figure}

\subsubsection{Training} 
We consider three variants with different model sizes, CMSA-S, -B, and -L, as shown in Table \ref{tab:arch_details}. 
Models are trained from scratch for 210 epochs with mean squared error loss, following the DeiT recipe \cite{touvron2021deit}. Training uses AdamW \cite{2019adamw} with linear warm-up for five epochs, an initial learning rate of $2.5 \times 10^{-3}$ (minimum $10^{-5}$), weight decay $5 \times 10^{-2}$, momentum $0.9$, and batch size 32.
Data augmentation includes random rotation, random scale, flipping, and random erasing. For comparison, we evaluate state-of-the-art methods using their official repositories\footnote{\textcolor{black}{UDP: https://github.com/HuangJunJie2017/UDP-Pose, \\HRFormer: https://github.com/HRNet/HRFormer, \\ ViTPose: https://github.com/ViTAE-Transformer/ViTPose, \\PCT: https://github.com/Gengzigang/PCT, 
}}. They are trained and evaluated on smaller images, which were not tested in the original papers.

\subsubsection{Results}
Table \ref{table:hum_pose} presents the results at various input resolutions. Following prior studies, we employ the standard evaluation metrics based on object keypoint similarity (OKS) to gauge the accuracy of the models. Our CMSAs demonstrate superior performance across all tested resolutions, notably surpassing competitors by a wider margin at lower input resolutions\footnote{It is noteworthy that they tend to show slightly lower AR scores, it is attributable to the choice of learning rate schedulers; when trained with the same step scheduler (\texttt{StepLRScheduler} in PyTorch) as others including ViT-B, the models (CMSA-B* and -L* in Table \ref{table:hum_pose}) yield superior AR scores.}. This underscores our model's efficacy in delivering high performance for low-resolution images. 

Figure~\ref{fig:ap_param_pose} presents the performance of methods using typical backbones with multi-scale representation for human pose estimation on $32\times 24$ images. Our CMSAs consistently outperform the comparison methods, demonstrating superior multi-scale representation abilities on low-resolution images.

\begin{table*}[ht]
  \centering
   \caption{Results on landmark-free head pose estimation with the BIWI and AFLW2000 datasets. Models are trained on the 300W-LP dataset. All FLOPs are calculated at single-scale. $^\dagger$ means the results reported in the respective papers. }
   \smallskip
   \begin{tabular}{c | c | c | c | c c c c| c c c c}
   \toprule
   \multirow{2}*{Model} & Input & Params &FLOPs & \multicolumn{4}{c|}{ BIWI } & \multicolumn{4}{c}{ AFLW2000} \\ 
   \cline{5-12} & Resolution&  (M)& (G)&Yaw & Pitch & Roll & MAE$\downarrow$  & Yaw & Pitch & Roll & MAE $\downarrow$\\ 
   \hline
   Hopenet($\alpha$=2) $^\dagger$ \cite{ruiz2018hopenet}  &224$\times$224& 23.9 & 4.1 & 5.17 & 6.98 & 3.39 & 5.18 & 6.47 & 6.56 & 5.44 & 6.16 \\
   Shao $^\dagger$ \textit{et al.} \cite{shao2019hpe}&224$\times$224 & 24.6 & 4.1 & 4.59& 7.25 & 6.15 & 6.00 & 5.07 & 6.37 & 4.99 & 5.48 \\
   WHENet $^\dagger$ \cite{zhou2020whenet} &224$\times$224& 4.4 & 0.4 & 3.99 & 4.39 & 3.06 & 3.81  & 5.11 & 6.24 & 4.92 & 5.42\\
   TokenHPE-v2 $^\dagger$  \cite{zhang2023tokenhpe}&224$\times$224& 86.4& 17.0 & 3.95& 4.51 & 2.71 & 3.72  & 4.36 & 5.54 & 4.08 & 4.66\\
   FDN $^\dagger$ \cite{zhang2020fdn}  &224$\times$224& -&- & 4.52 & 4.70 & \textbf{2.56} & 3.93  & 3.78 & 5.61 & 3.88 & 4.42\\
    Li $^\dagger$ \textit{et al.} \cite{li2022accurate}&224$\times$224& 0.2& - & \textbf{3.59} & \textbf{3.94}& 2.68& \textbf{3.40}& \textbf{3.36} & 5.05 & 3.56 & 3.99\\
   EVA-GCN $^\dagger$ \cite{xin2021eva}& 256$\times$256&$\sim$3.3&-& 4.46 & 5.34 & 4.11 & 4.64  & 4.01 & \textbf{4.78} & \textbf{2.98} & \textbf{3.92}\\
   MFDNet $^\dagger$ \cite{liu2021mfdnet}&224$\times$224&-&-& 3.40 & 4.68 & 2.77 & 3.62 & 4.30 & 5.16 & 3.69 & 4.38\\
   6DoF-HPE $^\dagger$  \cite{algabri20246DofHPE}& 224$\times$224& 37.9& 8.2& 3.91 & 4.43 & 2.69 & 3.68 & 3.56 & 4.74 & 3.35 & 3.88   \\
   6DRepNet $^\dagger$  \cite{hempel20246DRepNet360}& 224$\times$224& 43.8 & 11.2 &3.23 & 5.32 & 2.78 & 3.78 & 3.27 & 4.58 & 2.98 & 3.61   \\
   \hline
   Li $^\dagger$ \textit{et al.} \cite{li2022accurate}&56$\times$56& 0.2&- &4.85& 5.92& 3.29& 4.69& 4.48& 6.55& 4.80 & 5.28\\
   TokenHPE-v2  \cite{zhang2023tokenhpe}&64$\times$64& 85.9& 17.0 &5.79& 5.72 & 2.98 & 4.83  & 4.54 & 6.43 & 4.55 & 5.17 \\
   FSA-Net $^\dagger$ \cite{yang2019fsanet} &64$\times$64& 1.2 &-&4.27 & 4.96 & \textbf{2.76} & 4.00  & 4.50 & 6.08 & 4.64 & 5.07\\
   TriNet $^\dagger$  \cite{cao2021trinet}  &64$\times$64& 26.0&1.1 &\textbf{4.11} & 4.76 & 3.05 & \textbf{3.97} & \textbf{4.04} & 5.77 & 4.20 & 4.67 \\
   2DHeadPose \cite{wang20232dheadpose}&64$\times$64& 24.7&1.6 &4.42 &5.41 &3.37 &4.40 &4.17 &6.61 &4.59 &5.12 \\
   6DoF-HPE   \cite{algabri20246DofHPE}&64$\times$64&37.9&2.5 &4.56  & 4.79 & 2.80 & 4.05 & 4.80 & 6.12 & 4.32 & 5.08 \\
   6DRepNet   \cite{algabri20246DofHPE}&64$\times$64&43.8&2.9 & 4.56 & 4.64 & 2.85 & 4.02 & 4.89 & 5.87 & 4.38 & 5.04\\
   \rowcolor{mygray}
   \textbf{CMSA-B}  &64$\times$64 & 5.4&4.7 & 4.47& 5.29& 3.15& 4.30& 4.08& 5.30& 4.06& 4.48\\
   \rowcolor{mygray}
   \textbf{CMSA-L}  &64$\times$64 & 7.1& 5.8&4.49& \textbf{4.63}& 2.85& 3.99& 4.15& \textbf{5.10}& \textbf{3.89}& \textbf{4.38}\\
   \hline
   TriNet  \cite{cao2021trinet} &32$\times$32& 26.0&1.1& 4.96& 7.02& 3.75& 5.24& 6.79& 9.10& 7.54& 7.81\\
   Li $^\dagger$ \textit{et al.} \cite{li2022accurate}&28$\times$28& 0.2& - & 6.49& 9.12& 4.71& 6.77& 6.52& 8.56& 7.19 & 7.42\\
   TokenHPE-v2  \cite{zhang2023tokenhpe}&32$\times$32&85.7&16.9 & 6.83 & 6.41 & 3.76 & 5.66 & 5.35 & 7.34 & 5.46 &6.05\\
   2DHeadPose \cite{wang20232dheadpose}&32$\times$32&24.7 &1.5 &6.06 &7.49 &3.73 &5.76 &5.14 &7.55 &5.61 & 6.10\\
   6DoF-HPE  \cite{algabri20246DofHPE}&32$\times$32&37.9& 2.4& 5.78 & 7.80 & 3.56 & 5.71 & 6.23 & 6.72 & 5.08 & 6.01   \\
   6DRepNet  \cite{algabri20246DofHPE}&32$\times$32&43.8 &2.8 &6.69 &6.14 &3.69 &5.51 &6.37 &6.85 &4.98 &6.07   \\
   \rowcolor{mygray}
   \textbf{CMSA-S}  &32$\times$32 & 4.0&0.9 &\textbf{4.81}& \textbf{4.85}& 3.24& \textbf{4.33}& 4.01& 5.65& 4.12& 4.59\\
   \rowcolor{mygray}
   \textbf{CMSA-B}  &32$\times$32 & 5.4&1.2 &5.28& 5.25& 3.13& 4.55& 3.88& 5.58& 4.07& 4.51\\
   \rowcolor{mygray}
   \textbf{CMSA-L}  &32$\times$32 & 7.1&1.5 &4.97& 5.36& \textbf{3.12}&  4.48& \textbf{3.84}& \textbf{5.54}& \textbf{3.99}& \textbf{4.46}\\
   \bottomrule
  \end{tabular}
  \label{tab:head_pose}
\end{table*}

\subsection{Head Pose Estimation}
\subsubsection{Datasets}

We apply our model to head pose estimation without relying on facial landmarks, where models predict pitch, roll, and yaw angles directly from cropped images. Before being input to the models, these images are resized to a standard resolution. We adjust this resolution to manage the size of the input images effectively.
Models are trained on 300W-LP \cite{zhu2016aflw2000} and tested on BIWI \cite{fanelli2013biwi} and AFLW2000 \cite{zhu2016aflw2000}. 
300W-LP dataset contains 61,225 images (122,450 samples with flipping) covering large poses  ([$-90^{\circ}$, $90^{\circ}$]),
while BIWI has roughly 15,000 frames with small angles (yaw: $\pm75^{\circ}$, pitch: $\pm60^{\circ}$, roll: $\pm50^{\circ}$), generated from RGB-D videos, captured by a Kinect device for different subjects and head poses. Most of the faces in the BIWI dataset are small angles (yaw: $\pm75^{\circ}$, pitch: $\pm60^{\circ}$, roll: $\pm50^{\circ}$), while 300W-LP extends the original 300-W dataset to a large pose ([$-90^{\circ}$, $90^{\circ}$]).
AFLW2000 provides 2,000 images with 3D annotations and 68 landmarks. 
Following prior work \cite{ruiz2018hopenet, yang2019fsanet}, we discard 31 images with angles outside of [$-99^{\circ}$, $99^{\circ}$] in AFLW2000. 

\subsubsection{Training} 
We apply binned classification and soft stage-wise regression to estimate head pose from a single RGB image. Models are trained for 100 epochs on the 300W-LP dataset following the recipe from DeiT \cite{touvron2021deit}, with AdamW, cosine learning rate scheduling, and an initial learning rate $2.5\times 10^{-3}$. 
Input images are cropped to include the full head and normalized with ImageNet statistics \cite{deng2009imagenet}, using a batch size of 96. We use the official repositories for baselines\footnote{ \textcolor{black} {FSA-Net: https://github.com/shamangary/FSA-Net,  } \\ {TriNet: https://github.com/anArkitek/TriNet\_WACV2021,} \\ {TokenHPE-V2: https://github.com/zc2023/TokenHPE,} \\ {6DoF-HPE: https://github.com/Redhwan-A/6DoFHPE,} \\{6DRepNet: https://github.com/thohemp/6DRepNet360.} \\}.

\subsubsection{Results}
Table \ref{tab:head_pose} shows the results of ours and the state-of-the-art methods, where the standard evaluation metric, the mean absolute error (MAE), is adopted. At low-resolution settings of $64\times 64$ and $32\times 32$, our CMSAs can obtain state-of-the-art performance for the AFLW2000 dataset. Especially, CMSA-L achieves an MAE of 4.46 on $32\times 32$ images from AFLW2000, surpassing the performance of most typical models for higher image resolutions. For low-resolution images from BIWI dataset, our CMSA-L consistently achieves state-of-the-art performance.

\subsection{Image Classification with CIFAR-10/100}
\subsubsection{Training}
We train our CMSAs and existing models from scratch for 300 epochs on CIFAR-10 and CIFAR-100 at a resolution of $32\times 32$, following DeiT's \cite{touvron2021deit} recipe, with identical data augmentation, cross-entropy loss function, and AdamW optimizer. The initial learning rate was $2.5 \times 10^{-3}$ with a batch size of 128. 

\subsubsection{Results}
\begin{figure}[ht]
\begin{center}
\includegraphics[width=1.\linewidth]{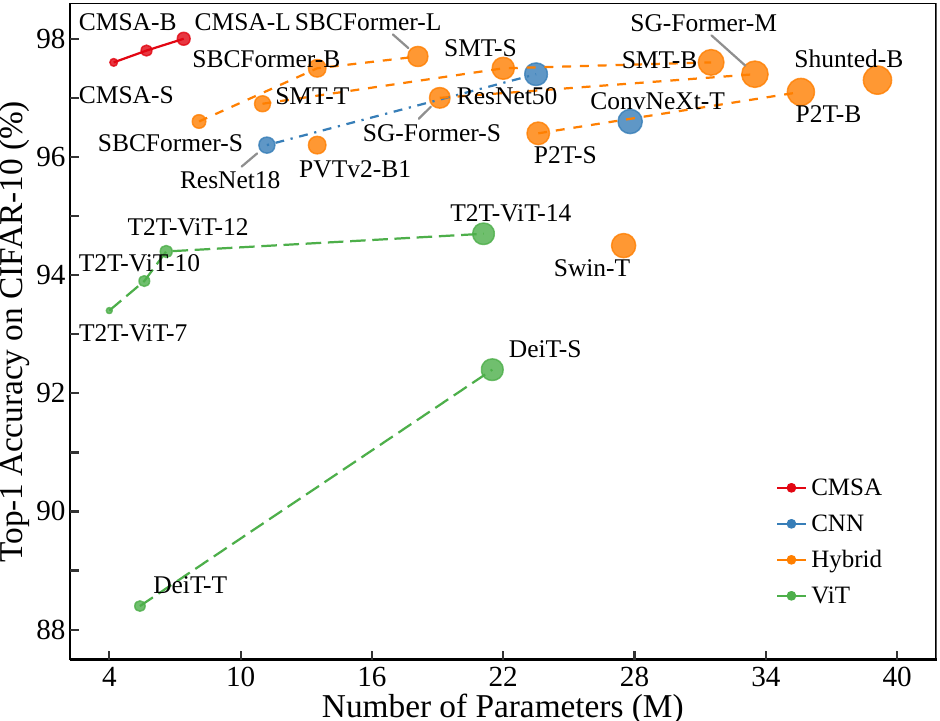}
\end{center}
\caption{Accuracy vs. Parameters. All models are trained for CIFAR-10 classification with input size $32\times 32$.}
\label{fig:acc_cifar10}
\end{figure}

Figure~\ref{fig:acc_cifar10} and Table \ref{tab:cifar} present the performance of our models across varying parameter sizes compared to representative methods in CNNs, ViTs, and their hybrids. It shows averaged values over 300 trials. Notably, our largest model, CMSA-L, achieves 98.0\% accuracy on CIFAR-10 and 85.2\% accuracy on CIFAR-100,  setting new benchmarks when limited to learning less than 300 epochs.

\begin{table}[ht]
    \centering
    \caption{Results on CIFAR-100 with input size $32\times 32$. $^\dagger$ means the values reported in the respective papers.} 
    \smallskip
    \begin{tabular}{c|c|c|c}
    \toprule
    \multirow{2}*{Model} & Params & FLOPs &CIFAR-100 \\
        & (M) & (G) &(\%)\\
    \hline
    DeiT-B $^\dagger$ \cite{touvron2021deit}  &85.1&21.8 &70.5  \\
    MobileViT-S \cite{mehta2022mobilevit}  &5.1&0.1 &75.2  \\
    T2T-ViT$_t$-14  \cite{yuan2021t2tvit}  &21.1&1.4 &75.9 \\
    Swin-B $^\dagger$ \cite{liu2021swin} &86.7& 5.0 & 78.5\\
    P2T-B  \cite{wu2022p2t} &35.6 &2.1 &79.6\\
    ConvNeXt-T  \cite{liu2022convnext}   &27.8  &1.5 & 80.3 \\
    PVTv2-B1  \cite{wang2022pvtv2}  &13.5&0.7 &80.5  \\
    P2T-L \cite{wu2022p2t} &53.9& 3.5 &81.8 \\
    ConvNeXt-B  \cite{liu2022convnext} &87.7&5.0  & 82.0 \\
    ResNet-50 \cite{he2016resnet} &23.5 &1.3 &82.9 \\
    SG-Former-M \cite{ren2023sgformer} &38.1&2.2 &83.1\\
    SMT-B \cite{lin2023scaleSMT} &31.6 &2.5 & 83.5\\
    Shunted-B \cite{ren2022shunted} &39.1&2.2 &83.5 \\
    SG-Former-B \cite{ren2023sgformer} &77.1&4.7 &83.8 \\
    SBCFormer-L  \cite{lu2024sbcformer} &18.2&3.2 & 84.1   \\
    \rowcolor{mygray}
    \textbf{CMSA-S} &4.2&0.9 &84.5  \\
    \rowcolor{mygray}
     \textbf{CMSA-B} &5.7&1.2 &85.0  \\
    \rowcolor{mygray}
    \textbf{CMSA-L} &7.4& 1.5 &\textbf{85.2}  \\
    \bottomrule
    \end{tabular}
    \label{tab:cifar}
\end{table}

\subsection{Semantic Segmentation}
\subsubsection{Training}
To evaluate CMSA on fine-grained perception tasks, we conduct semantic segmentation experiments on the Cityscapes \cite{cordts2016cityscapes} dataset using uniformly downsampled inputs of $64\times128$ resolution. ResNet-18 and ResNet-50 \cite{he2016resnet} are adopted as backbone networks with a fixed FCN segmentation head \cite{long2015FCNHead}. A single CMSA block with window sizes $\{(16\times32),(8\times16),(4\times8)\}$ is integrated into the backbone without modifying the segmentation head. All models are trained for 80k iterations with a total batch size of 8, following the standard Cityscapes training protocol \cite{cordts2016cityscapes}, and whole-image inference is used without sliding-window evaluation.
\subsubsection{Results}

\begin{table}[ht] 
\centering
\caption{ Semantic segmentation results on the Cityscapes dataset with $64\times 128$ inputs. CMSA is integrated into the backbone, while the FCN segmentation head is fixed across all models.}
\smallskip
\begin{tabular}{c c c}
\toprule
 Backbone & CMSA & mIoU(\%) $\uparrow$\\
  \hline
    ResNet-18  & \ding{55}  & 34.1  \\
    ResNet-18  & \checkmark  & 35.1 \\ 
   ResNet-50  & \ding{55}  & 36.8\\
   ResNet-50  & \checkmark  & 37.9 \\
\bottomrule
\end{tabular}
\label{tab:mmseg}
\end{table}

Table~\ref{tab:mmseg} presents the semantic segmentation results with and without CMSA for both ResNet-18 and ResNet-50 backbones on Citycapes with $64\times128$ inputs. CMSA improves mIoU from 34.1\% to 35.1\% on ResNet-18 and from 36.8\% to 37.9\% on ResNet-50, yielding gains of 1.0 and 1.1 percentage points, respectively. The consistent gains indicate that CMSA effectively facilitates contextual aggregation for dense, pixel-level semantic segmentation under limited spatial details.

\begin{figure}[ht]
\begin{center}
   \includegraphics[width=1.\linewidth]{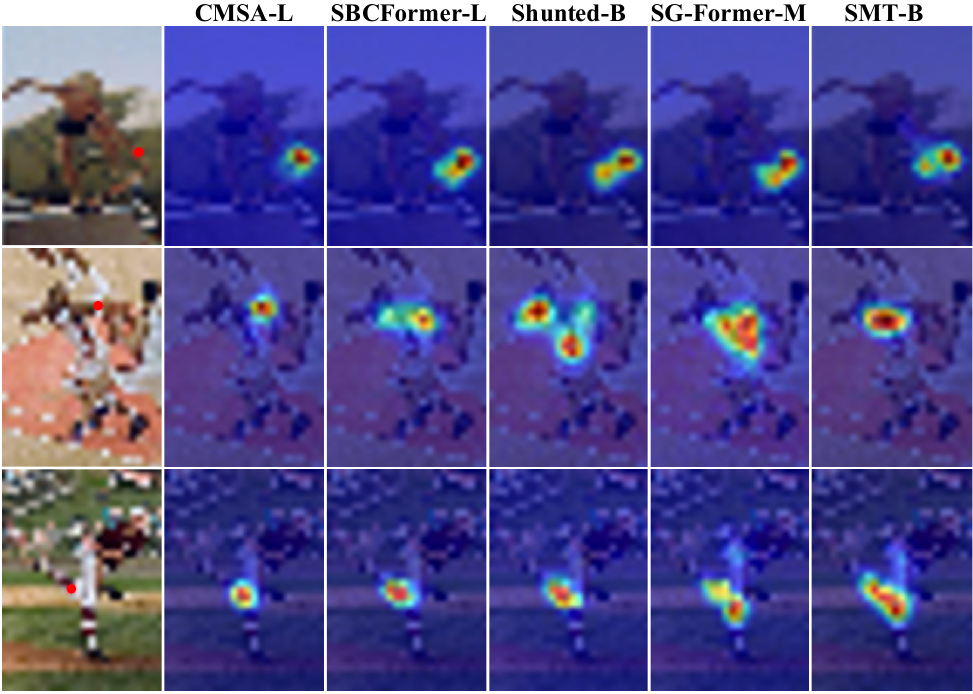}
   \end{center}
   \caption{Heatmap visualizations of the left wrist, elbow, and knee from models trained for human pose estimation with $32\times 24$ input. Redder regions indicate higher attention for the corresponding keypoint type in the heatmap.}
\label{fig:vis_heatmaps}
\end{figure}

\begin{figure}[ht]
\begin{center}
   \includegraphics[width=1.\linewidth]{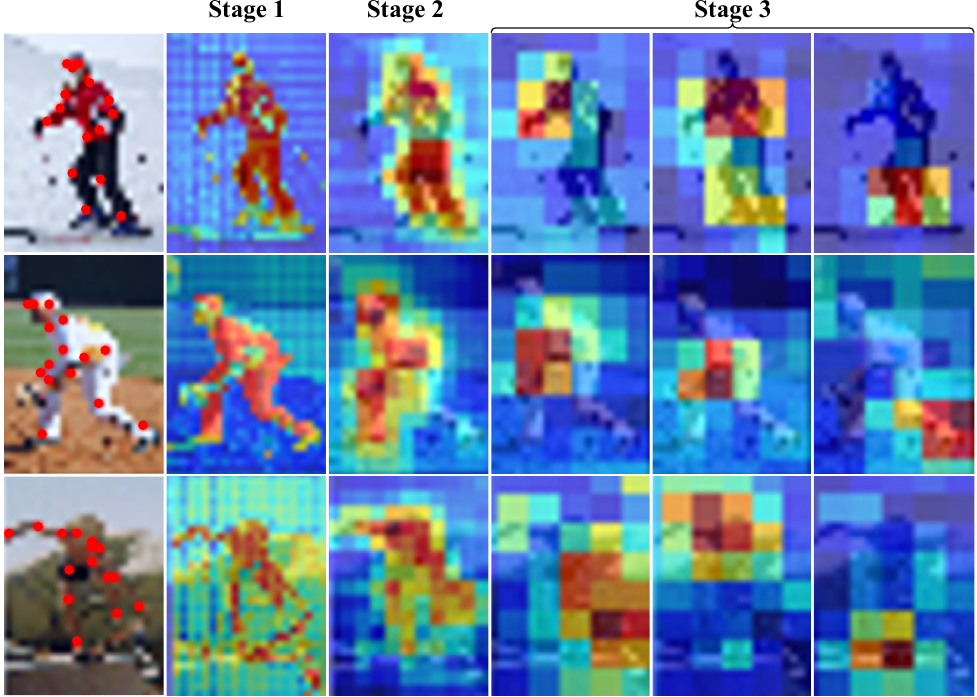}
   \end{center}
   \caption{Attention map visualizations from CMSA modules at each stage of CMSA-L for human pose estimation with input size $32\times 24$. Red points in the inputs denote the corresponding 17 keypoints.}
\label{fig:vis_attn_fea}
\end{figure}

\subsection{Visualization and Analysis}
To further validate the effectiveness of the proposed CMSAs, we visualize the predicted keypoint heatmaps generated by CMSA-L and methods using typical backbones with multi-scale representations for human pose estimation on the COCO 2017 dataset. As illustrated in Fig.~\ref{fig:vis_heatmaps}, our model produces more accurate and sharply focused activations around body keypoints compared to existing multi-scale attention methods. 
Furthermore, we visualize the attention maps from the final CMSA module of each stage in CMSA-L in Fig.~\ref{fig:vis_attn_fea}. The results indicate that, in the early stages, each keypoint attends broadly to other keypoints to capture a comprehensive global context. As the network deepens, the attention progressively focuses on smaller relevant regions, enhancing heatmap precision and keypoint localization. This contributes to the overall effectiveness of CMSA, particularly for low-resolution inputs.

\subsection{Ablation Study}

The proposed CMSA can be decomposed into the following components: grouped multi-scale attention, cascaded feature propagation, spatial fusion (SF), and channel fusion (CF).  
To assess the effectiveness of each, we conduct an ablation test with human pose estimation using COCO 2017.

\begin{table}[ht] 
\centering
\caption{Ablation tests of key components of CMSA for human pose estimation on COCO 2017 with $32\times 24$ inputs.}
\smallskip
\begin{tabular}{c c c c c c c c}
\toprule
 \multirow{2}*{Model} & Standard & Grouped &\multirow{2}*{Cascade} &\multirow{2}*{CF} &\multirow{2}*{SF} &\multirow{2}*{AP $\uparrow$}\\
 &  Attn & Attn & & & & \\
  \hline
    1  & \checkmark  & & & & & 48.3  \\
    2  & & \checkmark & & & &  51.8\\ 
   3  & & \checkmark &\checkmark & & & 53.5\\
   4  & & \checkmark & \checkmark& \checkmark & &53.7\\
   5  & & \checkmark & \checkmark&  & \checkmark &54.4\\
   6  & & \checkmark & \checkmark &\checkmark & \checkmark & 56.4\\
\bottomrule
\end{tabular}
\label{tab:ablation}
\end{table}

1) Effect of Each Component:
Table \ref{tab:ablation} presents the ablation results validating the contribution of each component in CMSA. The model in the first row, which has a hierarchical pyramid structure identical to our proposed network as shown in Fig.~\ref{fig:cmsa}a but uses standard self-attention instead of CMSA, performs the worst. The second-row model, while maintaining the same architecture, incorporates grouped multi-scale attention but lacks cascaded feature propagation, resulting in moderate performance improvements over the first model. The lack of explicit cross-scale information flow restricts further gains.  
The third-row model incorporates cascaded feature propagation, which enables progressive interaction between hierarchical features and leads to noticeable performance improvement even without CF and SF. The fourth-row model further includes CF, which adaptively fuses channel-wise features across scales, reducing redundancy in the cascaded representations. CF alone provides only a modest performance gain but effectively complements SF. The fifth-row model adds SF to enhance the representational power by facilitating the aggregation of spatial information. Finally, the sixth-row model integrates both CF and SF, allowing complementary channel-wise and spatial fusion, resulting in more effective multi-scale feature aggregation and the best overall performance. The performance improvements in these models demonstrate the effectiveness of the added components.

\begin{table}[ht]
\setlength\tabcolsep{1mm}
\centering
\caption{Ablation of group number $n$ and window sizes ($s_k \times t_k$) in the CMSA stages evaluated on COCO 2017 human pose estimation with $32\times 24$ inputs.}
\smallskip
\begin{tabular}{lcccc}
\toprule
\multirow{2}*{Model} & Stage 1 & Stage 2 & Stage 3 & \multirow{2}*{AP $\uparrow$}  \\
 &  $n$,\{$s_k$$\times$$t_k$\} & $n$,\{$s_k$$\times$$t_k$\}& $n$,\{$s_k$$\times$$t_k$\}&  \\
\midrule
1 & 2,\{8$\times$6, 8$\times$6\}  & 2,\{8$\times$6, 8$\times$6\} &2,\{8$\times$6, 8$\times$6\}  & 52.6\\
2 & 2,\{8$\times$6, 8$\times$6\}  & 2,\{16$\times$12, 8$\times$6\} &2,\{8$\times$6, 8$\times$6\}  &53.9\\
3 & 2,\{16$\times$12, 8$\times$6\}  & 2,\{16$\times$12, 8$\times$6\} &2,\{8$\times$6, 8$\times$6\}  &54.4\\
4 & 2,\{32$\times$24, 16$\times$12\}& 2,\{16$\times$12, 8$\times$6\} &2,\{8$\times$6, 8$\times$6\}  &55.6\\
5 & 3,\{32$\times$24, 16$\times$12, 8$\times$6\}  & 2,\{16$\times$12, 8$\times$6\} &2,\{8$\times$6, 8$\times$6\}  & 56.4\\
\bottomrule
\end{tabular}
\label{tab:ablation_groups_windows}
\end{table}

2) Effect of Group Number and Window Sizes: 
Table~\ref{tab:ablation_groups_windows} investigates the effects of the group number $n$ and window sizes ($s_k \times t_k$) in CMSA on COCO 2017 human pose estimation with $32\times24$ inputs. The first row model employs uniform small windows across all stages and yields the lowest performance, while progressively enlarging window sizes in earlier stages consistently improves performance by enhancing global context modeling. The fifth row model further increases the group number to $n=3$ in the first stage, enabling multi-scale attention within the same stage and yielding the best performance. These results confirm that multi-scale window scaling and multi-group attention contribute to effective multi-scale representation learning under low-resolution settings.

\begin{figure}[ht]
\begin{center}
   \includegraphics[width=1.\linewidth]{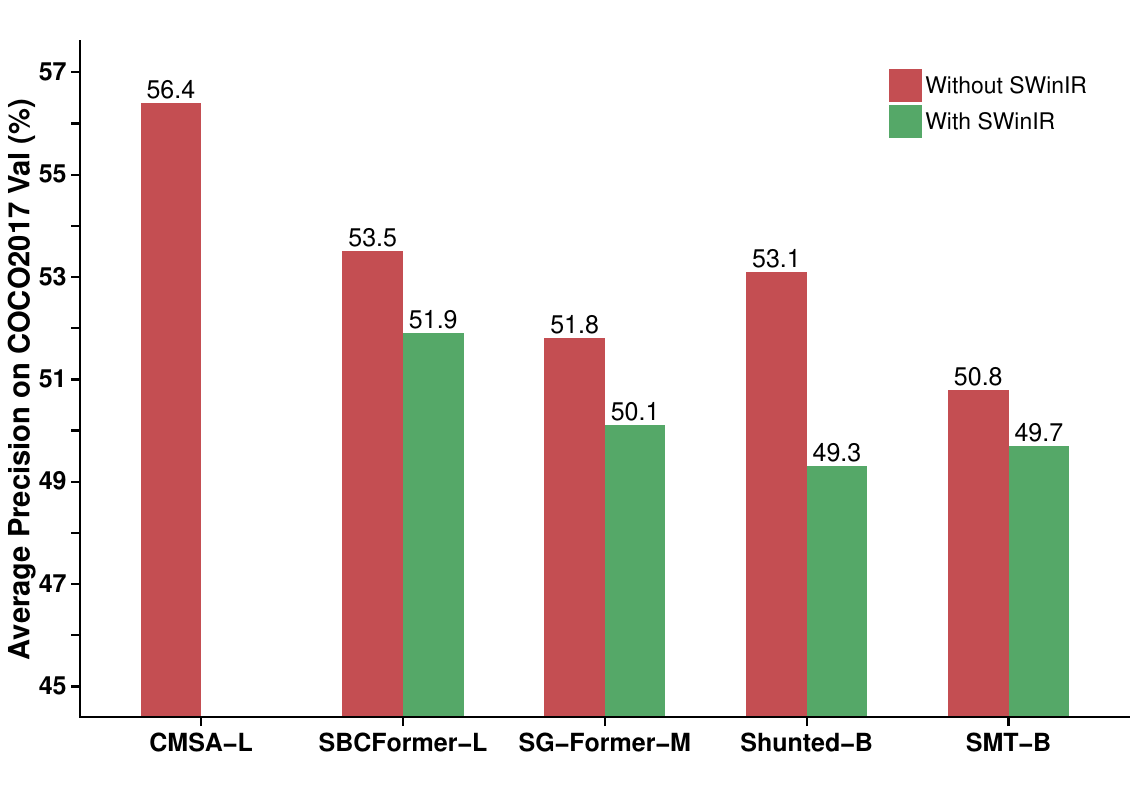}
   \end{center}
   \caption{Comparison of human pose estimation performance on COCO 2017 with $32\times24$ inputs, with and without super-resolution preprocessing, where low-resolution inputs are upsampled to $128\times96$ using SwinIR~\cite{liang2021swinir}.}
\label{fig:super_reso_enhance}
\end{figure}

3) Effect of Super-Resolution Preprocessing:
We evaluate the effect of super-resolution (SR) preprocessing by applying SwinIR \cite{liang2021swinir} to upsample $32\times24$ inputs to $128\times96$ prior to pose estimation. As illustrated in Fig.~\ref{fig:super_reso_enhance}, SR preprocessing reduces performance across different backbones for human pose estimation on the COCO 2017 dataset. This may be because SR preprocessing improves visual reconstruction but does not necessarily preserve the spatial cues required for accurate keypoint localization. These results highlight the effectiveness of CMSA in handling low-resolution inputs directly.

\section{Conclusion}
In this paper, we address the challenge of making visual inference on target objects from their low-resolution images, a critical issue in scenarios where targets are captured at low resolutions, such as with distant surveillance cameras or on low-performance edge devices. To mitigate the limitations inherent in low-resolution inputs—specifically, the difficulty in managing multi-scale features without compromising valuable information—we developed a novel attention mechanism, termed cascaded multi-scale attention (CMSA). This mechanism leverages a combination of multi-scale feature extraction and feature interactions, sidestepping the conventional downsampling operations for obtaining multi-scale features that typically result in loss of information in low-resolution contexts.

The empirical validation of our method across three tasks, i.e., human pose estimation, head pose estimation, and image classification, showed that our method surpasses current state-of-the-art approaches tailored to these applications, achieving this with fewer parameters. A preliminary semantic segmentation evaluation on the Cityscapes dataset also shows consistent improvements with CMSA. These results underscore the potential of our proposed method to significantly improve image recognition tasks under constrained conditions, marking a step forward in the deployment of deep learning models on edge devices and in situations where only low-resolution images are available. 
In conclusion, our research contributes a useful framework for enhancing the accuracy of image recognition in less-than-ideal conditions where high-resolution data is not accessible. Future work will explore its application to more challenging dense prediction tasks, such as small object detection and remote sensing, to further evaluate its scalability and effectiveness in complex visual environments.

\section*{Acknowledgments}
This work was supported by JSPS KAKENHI Grant Numbers JP23H00482.


\bibliographystyle{IEEEtran}
\bibliography{main}

\vspace{-1.4cm}
\begin{IEEEbiography}[{\includegraphics[width=1.in, height=1.15in, clip, keepaspectratio,trim={0cm} {0cm} {0cm} {0cm}]{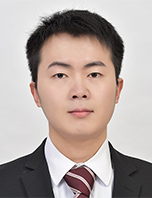}}]{Dr. Xiangyong Lu} received the Ph.D. degree in the Computer Vision Lab, Graduate School of Information Sciences, Tohoku University, under the guidance of Prof. Takayuki Okatani. He received the M.S. degree in the Service Computing Technology and System Lab, School of Computer Science and Technology, Huazhong University of Science and Technology in 2017. His research interests include computer vision, embodied AI, pervasive computing. 
\end{IEEEbiography}

\vspace{-1.4cm}
\begin{IEEEbiography}[{\includegraphics[width=1in,height=1.15in,clip,keepaspectratio,trim={0cm} {0cm} {0cm} {0cm}]{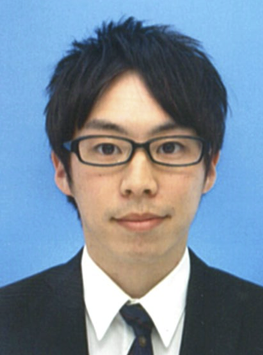}}]{Dr. Masanori Suganuma}
received the Ph.D. degree from Graduate School of Environment and Information Sciences, Yokohama National University, in 2017. He was an Assistant Professor at Tohoku University and currently is a research scientist at Sakana AI in Japan. His research interests are in the field of computer vision and machine learning.
\end{IEEEbiography}

\vspace{-1.4cm}
\begin{IEEEbiography}[{\includegraphics[width=1in,height=1.15in,clip,keepaspectratio,trim={0.5cm} {0.5cm} {0.5cm} {0.5cm}]{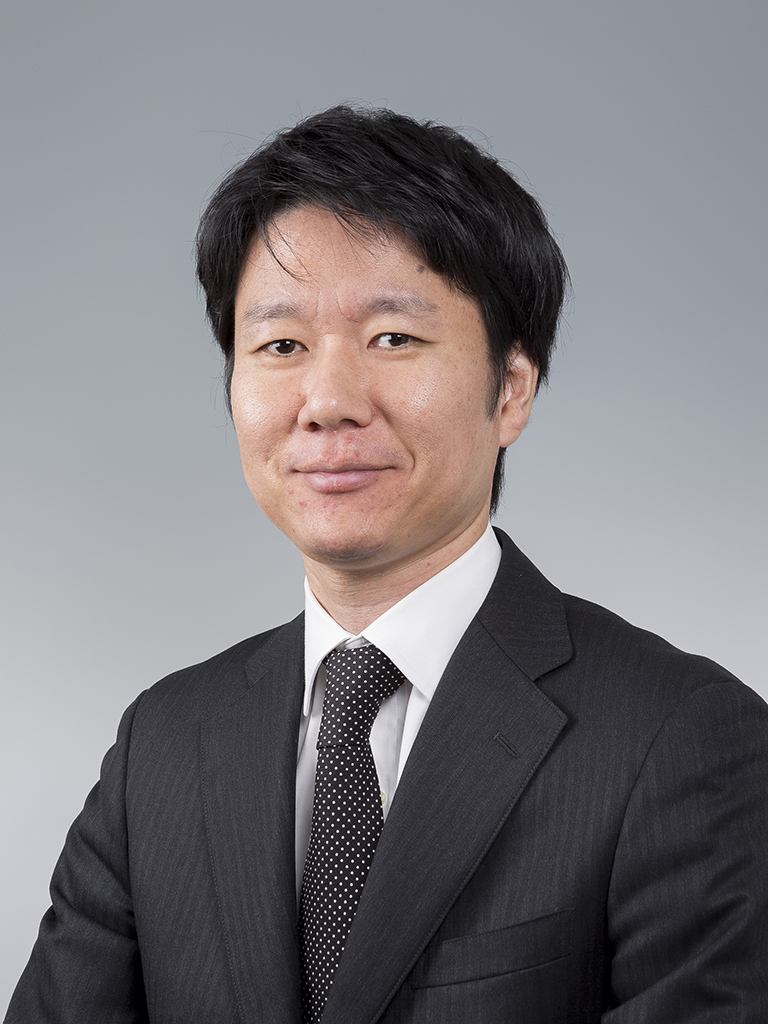}}]%
{Dr. Takayuki Okatani} earned his B.Eng., M.Sc., and Ph.D. degrees in Mathematical Engineering and Information Physics from the Graduate School of Engineering at the University of Tokyo in 1994, 1996, and 1999, respectively. He currently serves as a Professor in the area of computer vision at Tohoku University. In addition, he heads the Infrastructure Management Robotics Team at the RIKEN Center for Advanced Intelligence Project. With over 100 publications in peer-reviewed journals and conference proceedings, his work encompasses computer vision, deep learning, and multi-modal AI. He is an active member of several professional societies, including the IEEE Computer Society, the Information Processing Society of Japan, the Institute of Electronics, Information and Communication Engineers, and the Society of Instrument and Control Engineers.
\end{IEEEbiography}

\end{document}